\documentclass[review,10pt,authoryear]{elsarticle}

\usepackage{graphicx}
\usepackage{amssymb}
\usepackage{float}
\usepackage[left=1in, right = 1in, top = 1in, bottom = 1in]{geometry}
\usepackage{multirow}
\usepackage{blindtext}
\usepackage{hyperref}
\usepackage{amsmath}

\usepackage[table]{xcolor}

\journal{Machine Learning with Applications}

\begin{document}

\begin{frontmatter}

\title{An Applied Deep Learning Approach for Estimating Soybean Relative Maturity from UAV Imagery to Aid Plant Breeding Decisions}

\author[1]{Saba Moeinizade}
\author[2]{Hieu Pham}
\ead{hieu.pham@uah.edu}
\author[3]{Ye Han}
\author[3]{Austin Dobbels}
\author[1]{Guiping Hu}

\address[1]{Industrial and Manufacturing Systems Engineering, Iowa State University, Ames, IA 50011, USA}
\address[2]{College of Business, The University of Alabama in Huntsville, Huntsville, AL, 35899, USA}
\address[3]{Syngenta, Slater, IA 50244, USA}

\begin{abstract}
For a global breeding organization, identifying the next generation of superior crops is vital for its success. Recognizing new genetic varieties requires years of in-field testing to gather data about the crop's yield, pest resistance, heat resistance, etc. At the conclusion of the growing season, organizations need to determine which  varieties will be advanced to the next growing season (or sold to farmers) and which ones will be discarded from the candidate pool. Specifically for soybeans, identifying their relative maturity is a vital piece of information used for advancement decisions. However, this trait needs to be physically observed, and there are resource limitations (time, money, etc.) that bottleneck the data collection process.  To combat this, breeding organizations are moving toward advanced image capturing devices. In this paper, we develop a robust and automatic approach for estimating the relative maturity of soybeans using a time series of UAV images. An end-to-end hybrid model combining Convolutional Neural Networks (CNN) and Long Short-Term Memory (LSTM) is proposed to extract features and capture the sequential behavior of time series data. The proposed deep learning model was tested on six different environments across the United States. Results suggest the effectiveness of our proposed CNN-LSTM model compared to the local regression method. Furthermore, we demonstrate how this newfound information can be used to aid in plant breeding advancement decisions.

\end{abstract}


\end{frontmatter}


\section{Introduction}
Agriculture has been one of the most critical industries since human civilization. From manual-powered efforts thousands of years ago to the modern large-scale machine-driven farming, this industry is evolving to feed the increasing population while facing limited resources.  In agriculture, soybeans (\textit{Glycine max}) rank among one of the most important crops worldwide \citep{lee2015soybean, pagano2016importance}. Soybeans are not only a key source of protein for humans and livestock but also a key ingredient for biodiesel and vegetable oils. In 2019, the United States soybean exports totaled 19 billion USD with the global soybean market estimated to be over 50 billion USD each year since 2015 \citep{USDA2019Soy}.

Given the economic impact of soybeans, breeding companies, both in the private and public sector, endeavor to release improved soybean varieties to meet market demands where these new varieties are hopefully higher yielding than the previous generation. The process to identify superior varieties is through years of observed field trials. Within all soybean breeding pipelines, the relative maturity of a genetic soybean variety is an important characteristic to determine optimal growing environments, based mostly on latitude \citep{ortel2020soybean}. Identifying a soybean's relative maturity enables plant breeders to make optimal planting placement decisions to ensure a soybean reaches its maximum growth potential. That is, placing a genetic soybean variety in the correct latitude band will allow for the proper accumulation of sunlight needed for optimal growth. If a soybean variety is not placed into its optimal growing environment, the soybean may freeze before harvest, ultimately wasting precious resources and its yield potential. For a commercial breeding organization, at the end of each growing season, soybean advancement decisions must be made to determine which soybeans will be chosen to progress towards another year of field testing and which ones will be discarded. These meetings are typically considered the backbone of any breeding organization, and a company's success depends on identifying the best crops to move forward. It is common for these decisions to encompass 40,000 genetically different soybean varieties, and having key information is the difference between a confident decision and an educated guess. Soybean relative maturity is one of the most important factors (in addition to yield) in determining if a soybean is advanced or not. Therefore, it is vital that this information is accurately and efficiently collected before these decisions need to be made.

To determine the relative maturity of a soybean variety, soybean field trials are manually screened approximately once or twice per week by soybean physiology experts to determine the day of the year that 95\% of soybean pods within a plot have reached their mature brown color \citep{fehr1977stages}. However, for a global breeding organization, manually screening soybean fields is labor-intensive, time-consuming, and prone to human error, making the data collection process a difficult task. Of the 40,000 genetically different soybean varieties, only a fraction will have their relative maturity identified, leading to less optimal advancement decisions.

In recent years, digital and data-driven breeding/farming has led agriculture to a new era \citep{yost2011decision,aubert2012enabler,kurkalova2017sustainable,karimzadeh2019data,shahhosseini2020improved,moeinizade2019LAS,MoeinizadeMTLAS,moeinizade2021TI,Gorkem2020land,amini2021look, han2021dynamic}. Satellite imaging, live GPS tracking, drone-based fertilizing and screening, micro-sensors in the soil, etc., are all examples of new techniques that have brought significant changes to the agricultural industry. Modern image capturing devices such as unmanned aerial vehicles (UAVs)  allow researchers to gather images of crop fields at a faster rate with reduced dependency on error-prone, manual labor. Combined with modern analytical techniques, these images provide crop level data that breeders and farmers can use to enhance decision making \citep{nguyen2020monitoring, peng2020evaluation, khaki2020yieldnet}. A scalable image capturing technology has the ability to increase information throughput as well as reduce human error. However, the adoption of a new process leads to a new set of challenges, such as when and how frequently images should be collected and how algorithms can be trained to best convert images into useful data for decision making.  Oftentimes, the turnaround from the last drone flight to the advancement decision deadline is a couple of days. Therefore, any established data pipeline needs to be scalable to provide efficient data turnaround that is accurate for end goal needs.

Within machine learning, deep neural networks (DNN) are commonly used models that contain many sequentially stacked layers allowing the model to learn features that encode information from the image. Using a DNN, features are automatically learned from input data without manual feature creation. With increased computing power, automating image analysis is possible for agricultural tasks such as image classification, object detection, and object counting. These approaches often rely on a special type of DNN called convolutional neural networks (CNN). CNNs excel at image-based tasks because they take advantage of the spatial structure of the pixels. Additionally, CNNs require a fewer number of learned parameters, achieve superior computational performance compared to traditional machine learning, and reduce the risk of overfitting \citep{o2015introduction}. Given the flexibility and strength of CNNs, it comes as no surprise the vast amount of domains and applications in which they are used \citep{heinrich2021process, liu2021survey,tang2021model}.

Cornerstone deep learning frameworks such as RetinaNet \citep{alon2019tree}, Mask-RCNN \citep{yu2019fruit}, and YOLO \citep{mosley2020image} have been applied to detect sorghum heads, count strawberries, and classify tree species. Also given that commercial corn is a staple crop around the world due to being used for animal feed and biofuels, it is no surprise that there is substantial work combining corn images and deep learning \citep{khaki2020high, khaki2020convolutional, khaki2021deepcorn}. Aside from deep learning, many other researchers have combined traditional statistical modeling, machine learning, and image processing with agricultural data \citep{singh2016machine,naik2017real,moeinizade2018stochastic, dobbels2019soybean,moeinizade2020complementarity, pothen2020detection,LI2020304, Austin,shahhosseini2019maize, shahhosseini2020forecasting,shahhosseini2021coupling}. With the rapid combination of analytical techniques and plant breeding, this large body of recent work illustrate that large-scale analysis of imagery for improved decision making is possible.

Soybean relative maturity is an excellent example of a trait that can be measured using image analysis techniques.  Recent efforts have used drone-based imagery to estimate the days to soybean maturity, that is, the day of the year in which 95\% of pods within a plot have turned to their mature brown color.  These approaches have relied on multiple linear regression \citep{christenson2016predicting}, LOESS regression \citep{Austin}, segmented regression \citep{narayanan2019improving}, and partial least squares regression \citep{zhou2019estimation}. Each of these methods relies on condensing each RGB image to a single value, such as the normalized difference vegetation index or the green leaf index. However, such an approach lends itself to information loss due to summarizing a complete image towards a single number. Moreover, to date, \cite{trevisan2020high} is the only work that has used a CNN to estimate soybean maturity dates. However, their work does not take into account the temporal relationship between drone flight dates and simply uses an image classification approach. Therefore any extension of the literature requires addressing the temporal relationship between drone images, such as by using a long short term memory (LSTM) recurrent neural network \citep{yu2019review}. Like a CNN, an LSTM is a special type of deep neural network that can process multiple data points and sequences. Thus, they naturally lend themselves to be used for time series analysis. Given that important decisions depend on the accuracy of these predictions, it is vital that any edge in prediction accuracy is obtained.


Given this information, our research aims to

\begin{enumerate}
    \item Create a temporal, image-based, deep learning model (CNN-LSTM) to estimate soybean maturity
    \item Determine how our model's accuracy changes under the differing temporal frequency of flights
    \item Create a data-driven support system that will inform plant breeders when sufficient UAV imagery is captured for a soybean variety advancement decision
\end{enumerate}

Ideally, throughout a soybean's growth cycle, a drone flight should be conducted each day. However, in practice, due to limited resources (time, labor, money, etc.) and varying weather conditions, drone images may only be taken at most once or twice a week. Therefore it is the goal of this paper to create a system so that commercial breeding organizations can make decisions on when to best conduct drone flights and identify when enough flights are taken for an accurate soybean maturity date estimation to, ultimately, aid in advancement decisions. Additionally, if a prediction is within $\pm$ two days of the actual maturity date, that is enough to make a confident advancement decision. 

\section{Methods}
\begin{figure}[!htb]
    \centering
    \includegraphics[trim={2.55in 2.5in 2.55in 2in},clip, width =0.75\paperwidth]{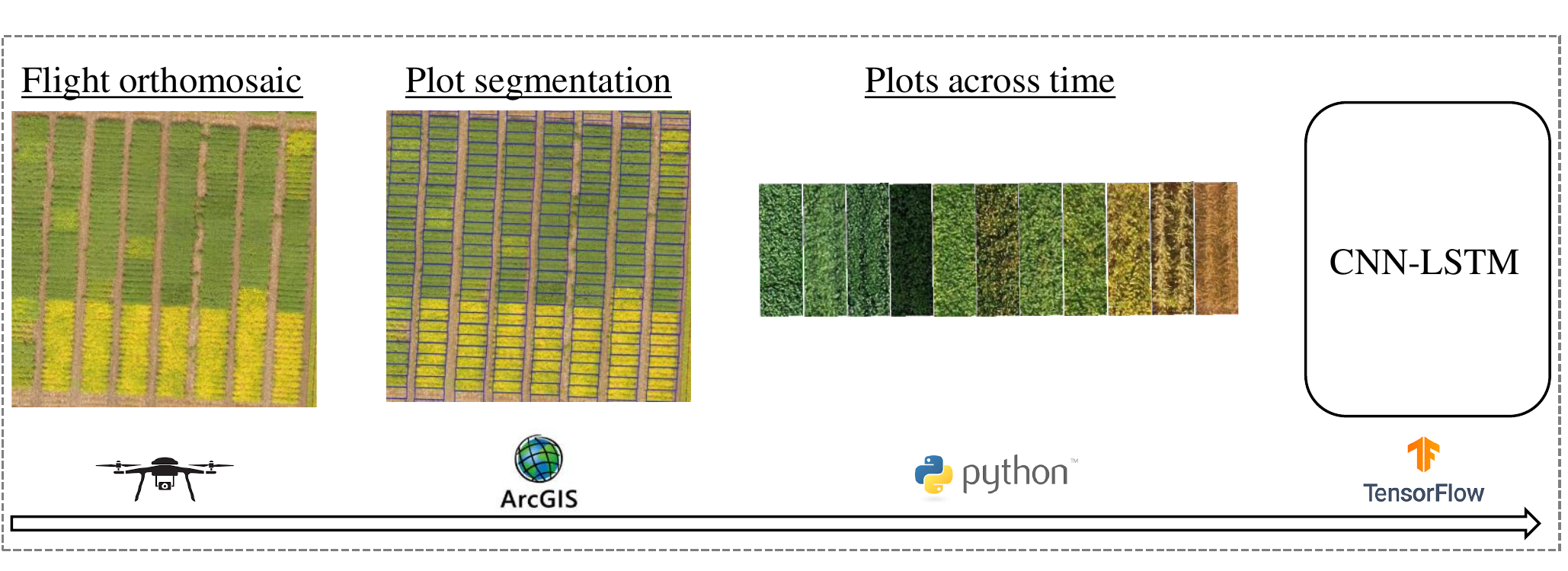}
    \caption{Illustration of the image capturing, data processing, and prediction process}
    \label{process}
\end{figure}

In this section, we introduce an end-to-end framework to systematically estimate the relative maturity of soybeans using computer vision and deep learning techniques. Our proposed framework comprises of four phases which are the data extraction and assembly, data pre-processing, feature extraction, and prediction phases. The data extraction and assembly phase first extracts plot images from each flight's full field orthomosaic and then assembles the time series of plot images for a given date. That is, each variety is snipped into a single image and then sequenced together by drone flight date. The data pre-processing phase includes rotating and resizing images to make the data consistent. In the feature extraction phase, time distributed convolutional neural networks are designed to automatically extract features from the time series of images for each variety. Finally, long short-term memory recurrent neural networks are used to predict the relative maturity of soybeans given features extracted from the previous phase. Figure \ref{process} illustrates this process for a single flight orthomosaic (one drone flight on a single day) for numerous soybean varieties. At each stage, we make use of open-source software for our end-to-end process. In practice, this entire pipeline is completed in an efficient manner, so that plant breeders have sufficient data to make necessary soybean advancement decisions.

\subsection{Proposed deep learning model}
\begin{figure}[!htb]
    \centering
    \includegraphics[width = 0.6\paperwidth]{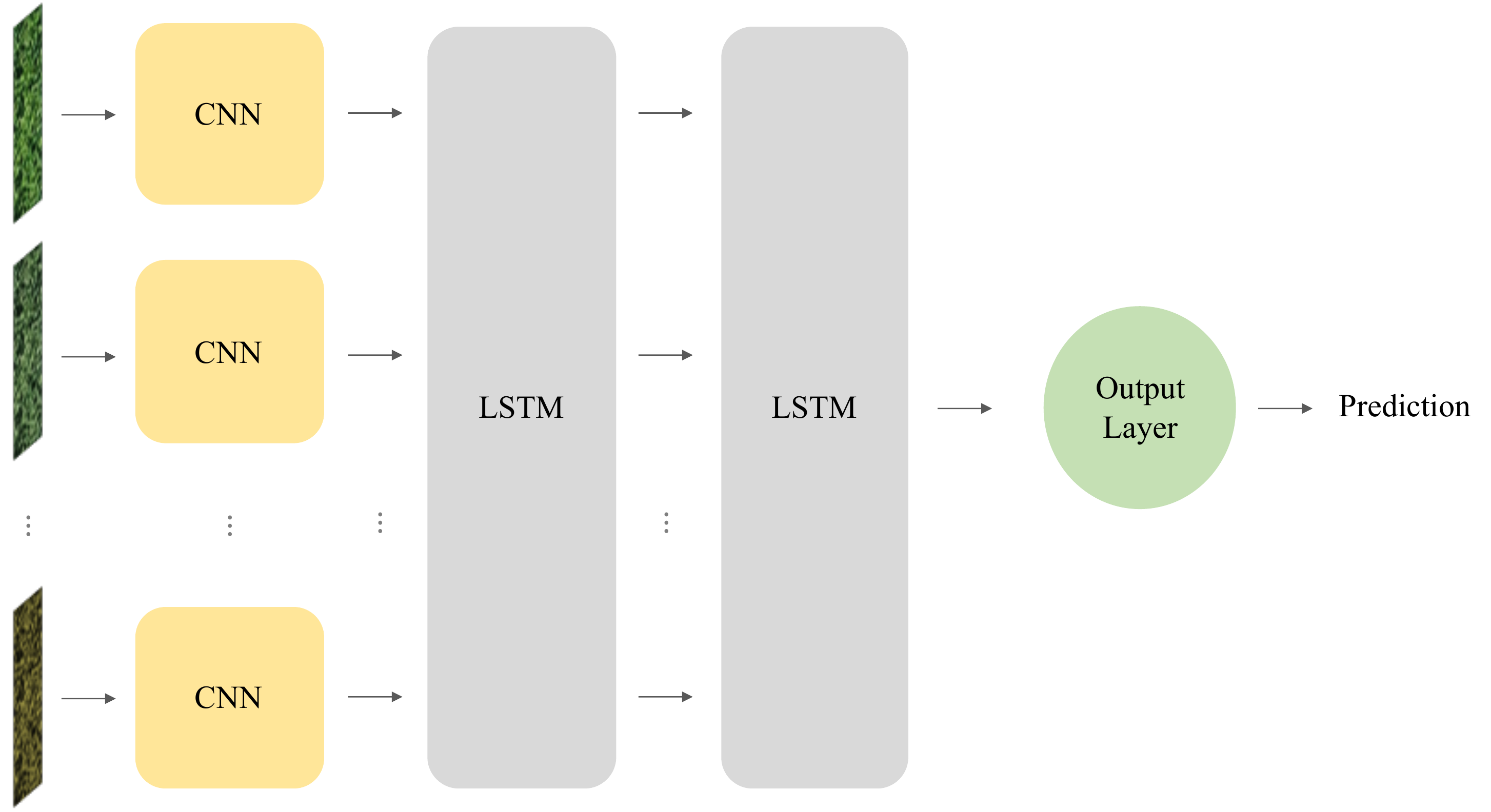}
    \caption{The CNN-LSTM structure used to predict soybean relative maturity from images.}
    \label{structure}
\end{figure}

Figure \ref{structure} demonstrates the structure of the deep learning model. In this hybrid model, for each plot snip, time-distributed CNNs are used for deep feature extraction, and LSTM is used for capturing the sequential behavior of time series data. Time distributed layers have the advantage of applying convolution on each image in the time series independently. 

In detail, each image is first passed to a set of convolutional and max pooling layers to produce a fixed-length vector representation. Convolutional layers act as a feature extractor that automatically learns specific details about the image. This contrasts with traditional machine learning methods, where features need to be handcrafted by a domain expert. Acting as our feature extractor, we have four convolutional layers and two max pooling layers. The repetitive nature of the convolutions helps to extract a more detailed structure about the images. As an illustration, the first couple of convolutional layers learn simple features such as colors and edges, but as the number of convolutions increase, the new features identify shading patterns, areas of interest, and pixel densities. At a high level, the max pooling layers act to reduce the dimensions from learned feature maps, and therefore reduce the number of parameters to learn, thus leading to less computational demand. In short, the convolutional layers act as a way to reduce an image down to a single vector of features so that the resulting output can be read into a learning algorithm.

\begin{figure}[!htb]
    \centering
    \includegraphics[width=\linewidth]{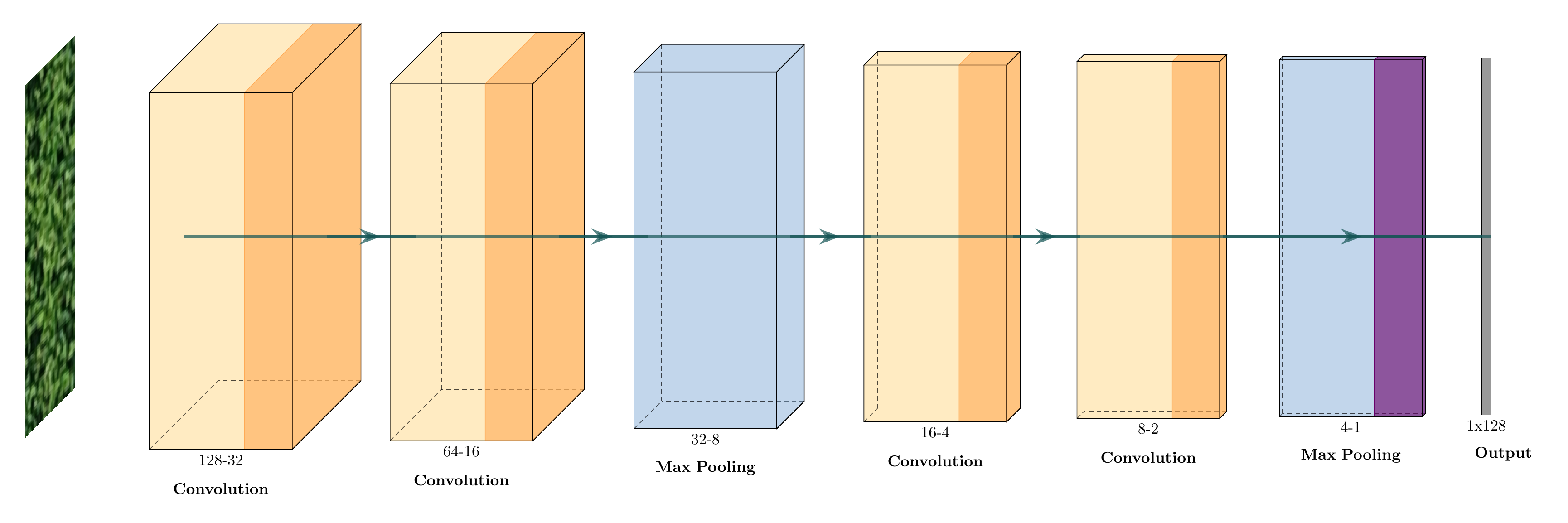}
    \caption{Illustration of the CNN architecture. The convolution layers apply convolution operation between the raw data and convolution kernels producing new features. The numerical values represent the dimensions of each layer (depth-width). The max pooling layers are down sampling the convolved features and help to make the system more robust since small changes of the input will not change the pooling output.}
    \label{CNNarchitecture}
\end{figure}

Figure \ref{CNNarchitecture} demonstrates the architecture of CNN layers (see section \ref{DOE} for more details about the network architecture). The outputs of CNN for the sequence of plot images are passed to an LSTM model. Long short-term memory (LSTM) is a special type of recurrent neural network (RNN) that is well suited for making predictions based on time series data. LSTMs solve the vanishing gradient problem that can be encountered in classical RNNs by using information gates to store useful information and forget unnecessary information. The output of the LSTM layer is passed to an output layer, and finally, relative maturity is estimated. More instructions about experiment design are presented in section \ref{DOE}.

\subsubsection{Loss Function}

To make the model more robust to outliers, we used Huber loss function \citep{huber1992robust} which is defined as follows:

\begin{equation}
L_{\delta}(y, \hat{y})=\left\{\begin{array}{ll}
\frac{1}{2}(y-\hat{y})^{2} & \text { for } \ |y-\hat{y}| \leq \delta \\
\delta|y-\hat{y}|-\frac{1}{2} \delta^{2} & \text { otherwise }
\end{array}\right.
\end{equation}

\noindent Here $y$ is the true label, and $\hat{y}$ is the prediction. For small values of $\delta$, this function is quadratic and for large values it is linear. Moreover, $\delta$ is a hyper-parameter that can be tuned. This loss function brings the advantages of mean square error (MSE) loss and mean absolute error (MAE) loss together by combining them in a piecewise manner.

\subsection{Benchmark}
We compare the performance of our proposed deep learning model against a recent state-of-the-art benchmark created by \cite{Austin} where a local regression model is fitted to the RGB color transformation values over time. That is, for each plot and drone flight date, an RGB transformation is performed, and the output is fed into the regression model. \cite{Austin} demonstrated that using the mean greenness leaf index (GLI) on each plot combined with locally estimated scatter plot smoothing (LOESS) regression results in higher correlations between the predicted and ground truth maturity days. The LOESS model combines much of the simplicity of linear least squares regression with the flexibility of nonlinear regression by fitting linear models to localized subsets of data determined by a nearest neighbor algorithm \citep{LOESScleveland1988locally}. In this study, the LOESS model was implemented using the ``lowess'' function from ``statmodels'' package in Python to interpolate the GLI values extracted from the time series of images. This function implements the algorithm described in \cite{LOESScleveland1979robust}. The GLI index can be calculated as follows:

\begin{equation}
    GLI= \frac{2G-R-B}{2G+R+B}
\end{equation}

\noindent where R, G, and B are the mean values of red, green, and blue channels respectively for each image. Finally, the relative maturity date is predicted as the day with the closest value to a predefined threshold, where the threshold value corresponds to a level of greenness. \cite{Austin} suggested that a GLI threshold of 0.02 was optimal from their extensive testing. However, this value can vary based on the environment and equipment, and optimizing a single threshold is not scalable. This is one limitation we hope to address with our CNN-LSTM approach.

\section{Experiments and Results}
In this section, we present the data sets used in this study, the deep learning model's hyper-parameter tuning, and the results for both the deep learning model and the LOESS model.

\subsection{Data}
Data from 6 different locations across the United States in two growing seasons (2018 and 2019) is used in this study. All data (orthomosaics, plot delineations, and ground-truth) except environment 3 (Elkhart, IA, 2019) were adopted from the University of Minnesota Soybean breeding project \citep{dataset}. The environment 3 data set is provided by a commercial breeding organization. The following table includes more details about the data sets used in this study. Environments 3 and 6 have the largest number of plots (soybean varieties), which is almost twice the number of plots in other environments.  During the growing season, a commercial breeding organization can have upwards of 40,000 genetically different soybeans planted in 200 geographically different locations (ranging from Mississippi to Canada). Due to the labor requirements and plant breeding expertise needed to accurately observe soybean maturity, it is challenging to have extensive ground-truth data. 

\begin{table}[ht]
    \centering
    \begin{tabular}{|c|c|c|c|c|c|}
    \hline
        Environment& Location & Year & Number of plots & Planting date & UAS Platform \\
        \hline
        1&  Waseca, MN & 2018 & 874    & May 17 & DJI Phantom 3 Pro\\
        2&  Lamberton, MN & 2018 & 796 & May 16 & DJI Phantom 3 Pro\\
        3&  Elkhart, IA & 2019 & 1686  & April 24 & DJI Phantom 3 Pro\\
        4&  Waseca, MN & 2019 & 688    & May 15 & DJI Phantom 3 Pro\\
        5&  Lamberton, MN & 2019 & 896 & May 16 & DJI Phantom 3 Pro\\
        6&  Rosemount, MN & 2019 & 1410& May 26 & DJI Inspire 1\\
         \hline
    \end{tabular}
    \caption{Data set descriptions. Each environment varies in number of plots, planting dates and unmanned aircraft system (UAS) platform.}
    \label{tab:datasets}
\end{table}

Each data set includes orthomosaics, plot boundary delineations, ground control points, and the ground-truth notes that were taken by soybean physiology experts visiting plots and manually observing the soybean's relative maturity. To extract individual plots from orthomosaic images, we used a Python script to parse the plot boundaries, and we tagged each image with the drone flight date and unique identifier (soybean's name). We performed this process for each orthomosaic across all flight dates as shown in Table \ref{tab:my_label}. In total, we processed 31,750 unique images. It should be noted that the number of plots is not exactly the same as the previous work conducted by \cite{Austin} since we kept outliers in the data sets to test the robustness of our proposed models. Whereas \cite{Austin} removed outliers. 

\begin{table}[]
    \centering
   
    \begin{tabular}{|c|c|}
     \hline
        Environment & Flight dates \\ 
        \hline
        1 & Sep 6, Sep 13, Sep 20, Sep 27, Oct 8 \\
        2 & Sep 5, Sep 14, Sep 17, Sep 25, Oct 4\\
        3 & Sep 7, Sep 13, Sep 20, Sep 26, Oct 3\\
        4 & Sep 6, Sep 14, Sep 21, Sep 27, Oct 7\\
        5 & Sep 6, Sep 14, Sep 18, Sep 25, Oct 4\\
        6 & Sep 6, Sep 13, Sep 20, Sep 27, Oct 7\\
        \hline
    \end{tabular}
    \caption{Flight dates across 6 environments. For each environment, 5 images are selected on a weekly basis starting from September till the first week of October. Bi-weekly images can be obtained by selecting the first, middle and last dates among these flights.}
    \label{tab:my_label}
\end{table}

\begin{figure}[H]
    \centering
    \includegraphics[width=\linewidth]{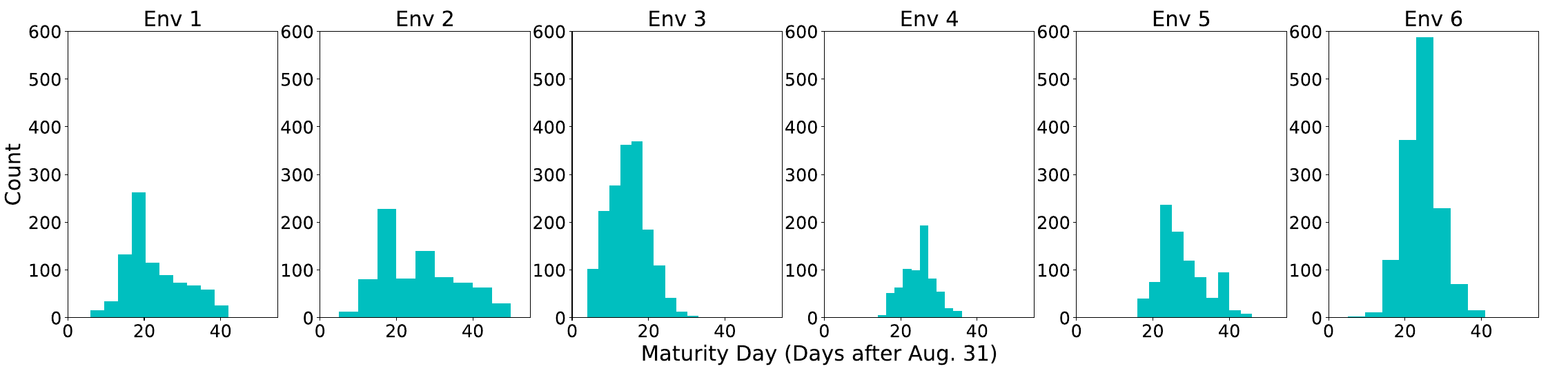}
    \caption{Histograms of relative maturity days for 6 environments. Relative maturity day is defined as days since August 31.}
    \label{fig:hist}
\end{figure}

The relative maturity days are calculated as the number of calendar days after August 31. That is, our model simply estimates a numeric value that we add to August 31 to get a month-day estimation for the maturity date. Figure \ref{fig:hist} visualizes the distribution of relative maturity days for each environment. As shown in Figure \ref{fig:hist} distributions vary in shape and spread. The median of relative maturity days across these 6 environments is 20, 24, 15, 25, 26, and 27, respectively. As expected, environment 3 (with the earliest planting date) has the lowest median, that is, the earliest maturity date.

All flights start from mid-August and last in mid-October, and each environment varies in flight dates. To be consistent, we selected weekly flights considering a 5-week period, including 4 weeks in September and the first week of October for each environment. We train our models on weekly and bi-weekly flights, including 5 and 3 images, respectively, to investigate the effect of less frequent flights on prediction performance. In practice, if accurate performance can be obtained from fewer flights, that would increase the speed of information gain, thus allowing sufficient time for plant breeders to make advancement decisions. Operationally, a reduced frequency in the number of drone flights would present significant cost savings both in time and reduce drone costs. In an ideal scenario, a breeding organization can fly drones every day, but due to restrictive factors (weather, time, labor, etc.), flights can only be conducted at most once or twice a week. A system that can yield accurate and confident results with fewer drone flights would have numerous positive outcomes for a breeding organization.

\subsection{Data Processing}
A number of pre-processing steps are applied to the images before moving on to the prediction
task. The first step is to resize all images to a fixed width and length. The original images of plots have a rectangular shape with width varying from 57 to 65 and length varying from 146 to 280. We tried different sizes and find out that the prediction is not very sensitive to the image size since we are mostly extracting features related to color and shape. Therefore, we decided to resize all images to a fixed size of $64 \times 256$ for consistency. 

Next, we applied some data augmentation techniques to test the robustness of the proposed deep learning model against some variation of images which can be caused due to cloudiness and the  relative position between the camera and sun. These techniques include changing the brightness and contrast and making some images blur. Data augmentation techniques have been conducted on 20\% of total images. To change the brightness of images, we added a constant to each pixel. Similarly, contrast can be changed by multiplying each pixel by a constant. To make the images blur, we used Gaussian smoothing to remove noise. 

\begin{figure}[H]
    \centering
    \includegraphics[width=10cm]{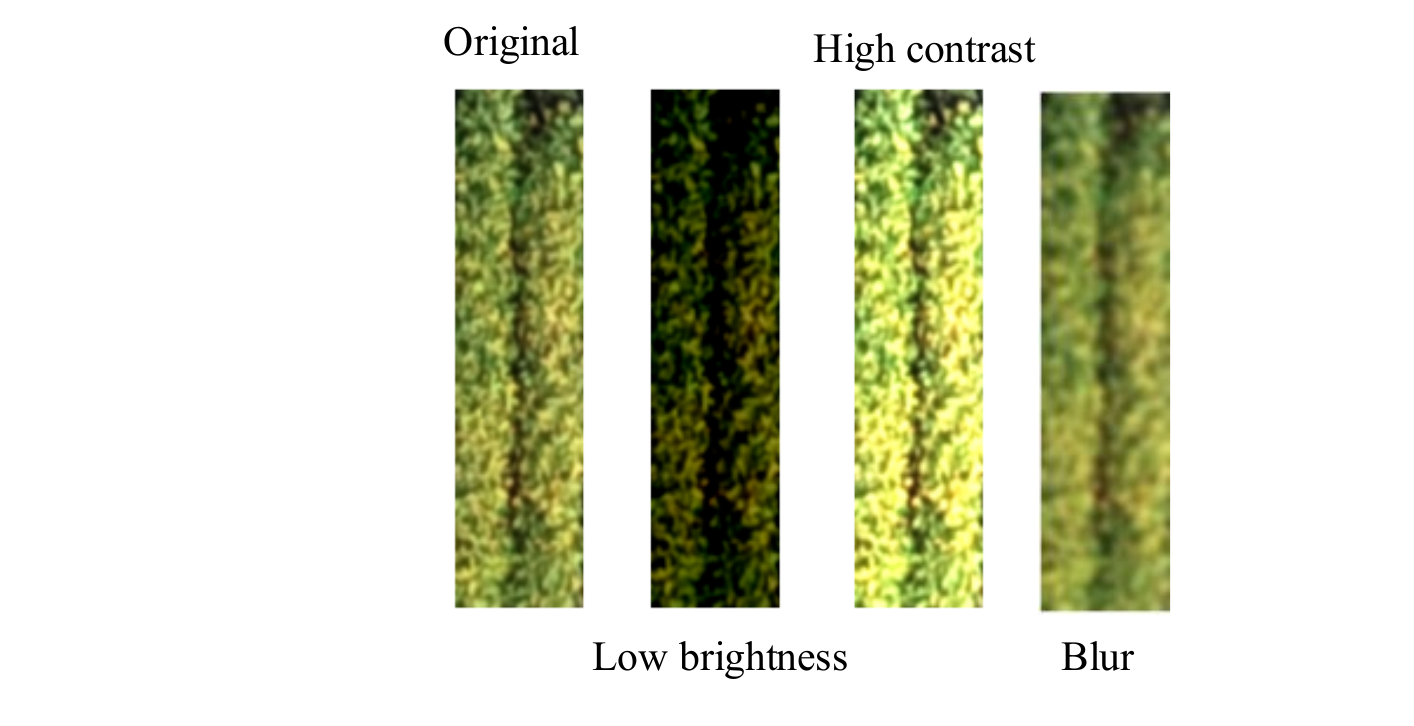}
    \caption{Representation of data augmentation for a single plot. To make an image dark, a negative constant (-100) is added to each pixel of the image (low brightness). High contrast image is obtained by multiplying each pixel by a constant (1.5) and the blur image is obtained by removing noise using guassian blur.}
    \label{CNrchitecture}
\end{figure}


\subsection{Design of Experiments} \label{DOE}
The CNN-LSTM model consists of four time-distributed convolutional layers of 32 filters and two LSTM layers. The detailed structure of the network is provided in Table \ref{tab:structure}. 
In the CNN model, downsampling is performed by max pooling with a stride of 2. The output of the last max pooling layer is flattened and used as input features for LSTM layers. The LSTM model has 256 hidden units. The first LSTM layer (LSTM$_{1}$) returns a sequence that goes as input to the second LSTM layer (LSTM$_{2}$) that outputs a vector. Finally, a dense layer with 1 neuron is applied to estimate the relative maturity of each plot. After trying different network designs, we found this architecture to provide the best overall performance. This architecture has a total of 948,449 trainable parameters. 

The weights of the network are initialized using Xavier initialization \citep{glorot2010understanding}, and activation functions are all rectified linear units (ReLU) except for the last layer, which is linear. The hyperparameter $\delta$ in the Huber loss function is set as 1. To determine this value, we applied a search grid over different values of $\delta$ and chose the optimal one. The loss function is optimized using stochastic gradient descent (SGD) with a batch size of 64 \citep{lecun1998gradient}. The optimization algorithm is Adaptive Moment Estimation (Adam) with a learning rate of $10^{-3}$ and decay rate of $10^{-1}$ \citep{kingma2014adam}. The model is trained for 300 iterations.

The proposed model was implemented in Python using the Keras library with the Tensorflow backend, and training time took about half an hour on an NVIDIA Tesla V100 SXM2 GPU.

\begin{table}[H]
    \centering
    \begin{tabular}{|c|c|c|c|c|c|}
    \hline
        \textbf{Layer name} & \textbf{Filter size} & \textbf{\# Filters} & \textbf{Stride} & \textbf{Padding} & \textbf{Input size}\\
        \hline
         Time distributed Conv2D$_1$ & $3\times 3$          & 32                & 2      & Same   & (5, 256, 64, 3)\\\hline
         Time distributed Conv2D$_2$ & $3\times 3$          & 32                & 2      & Same  & (5, 128, 32, 32) \\\hline
   Time distributed MaxPooling2D$_1$ & $2\times 2$         & -                & 2      & -   &(5, 64, 16, 32)\\\hline
         Time distributed Conv2D$_3$ & $3\times 3$         & 32                & 2      & Same  &(5, 32, 8, 32) \\\hline
         Time distributed Conv2D$_4$ & $3\times 3$         & 32                & 2      & Same  &(5, 16, 4, 32) \\\hline
   Time distributed MaxPooling2D$_2$ & $2\times 2$         & -                & 2      & -   &(5, 8, 2, 32)\\\hline
   Time distributed Flatten        & $3\times 3$        & -                & 2      & -   &(5, 4, 1, 32)\\\hline
                     LSTM$_1$        & $3\times 3$         & 32                & 2      & Same  &(5, 128) \\\hline
                     LSTM$_2$        & $3\times 3$        & 32                & 2      & Same   &(5, 256)\\\hline
                     Dense  (1 neuron)       &- &- &-& -  & 256\\ 
                     \hline
    \end{tabular}
    \caption{The full summary of our CNN-LSTM network.}
    \label{tab:structure}
\end{table}

According to Table \ref{tab:datasets} there exists total $N = 6350$ plots. We randomly select 85\% of plots as the input and use the rest as the test data to evaluate performance. Since the number of plots varies across different environments, we kept approximately 15\% of total plots per environment as test data. As such, the test size across the six environments is 117, 112, 260, 104, 134, and 226, respectively. The input data is split into train and validation sets randomly. The validation set (10 \% of input data) is used to monitor the training process.


\subsection{Performance Evaluation Metrics}
To measure the performance of the proposed model, we use the mean absolute error (MAE) and mean squared error (MSE) metrics, which are defined as follows:

\begin{equation}
MAE=\frac{1}{N} \sum_{i=1}^{N}\left|y_{i}^{pred}-y_{i}^{G T}\right|
\end{equation}

\begin{equation}
MSE=\frac{1}{N} \sum_{i=1}^{N}\left|y_{i}^{pred}-y_{i}^{G T}\right|^{2}
\end{equation}

\noindent Here $y_{i}^{G T}$ denotes the true relative maturity for the $i^{\text{th}}$ plot, $y_{i}^{pred}$ denotes the predicted relative maturity, and $N$ is the total number of plots.

\subsection{Results}
In this section, we provide the final results for the proposed CNN-LSTM model and the benchmark (LOESS model) and compare their performance.

\subsubsection{CNN-LSTM Performance}
\begin{figure}[!htb]
    \centering
    \includegraphics[width=12cm]{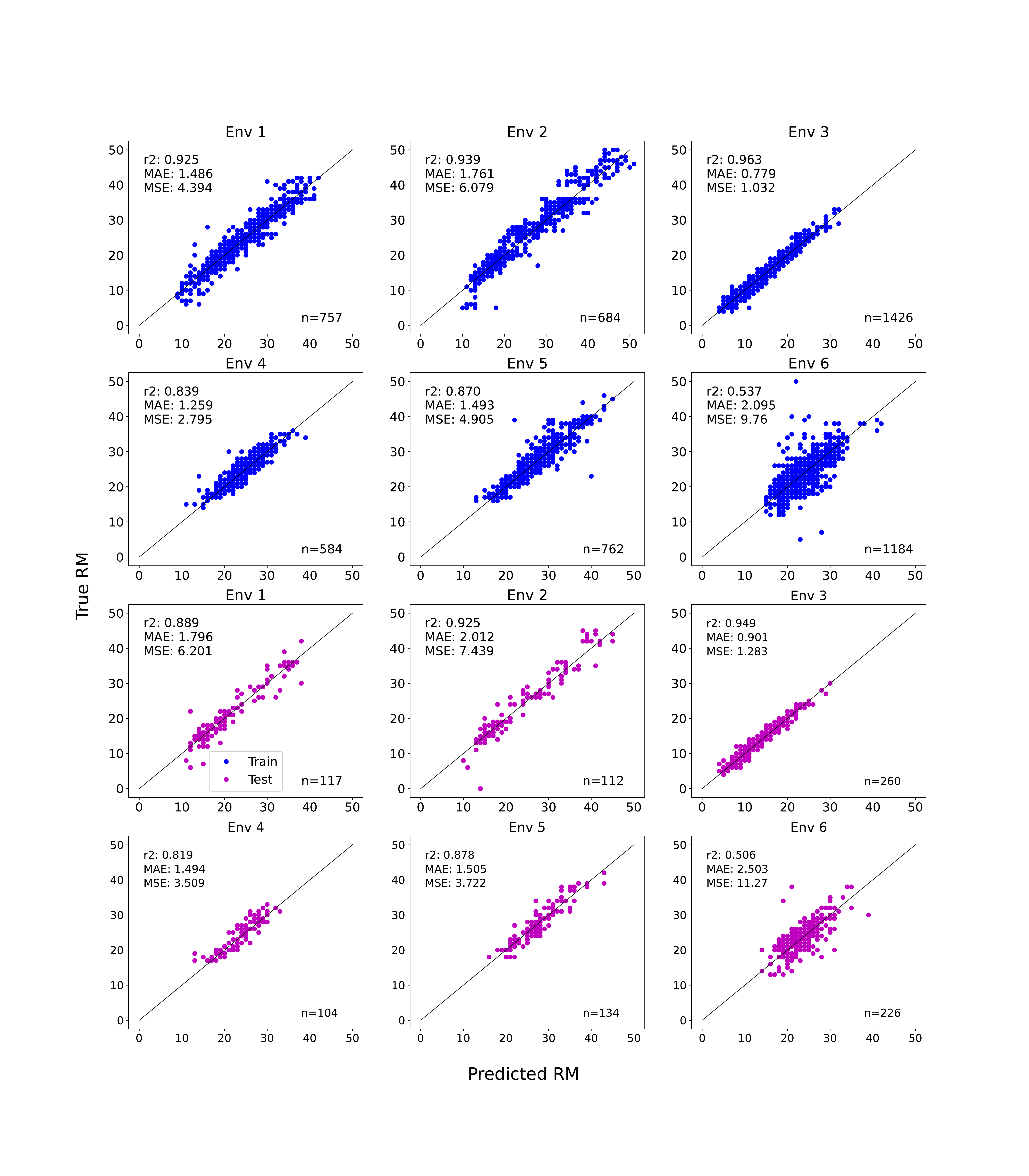}
    \caption{CNN-LSTM model's train (blue) and test (purple) performance across six environments for 5 images. Figures present the predicted RM versus ground truth. The black line corresponds to $x=y$. The performance metrics are presented in the upper left corner of figures and number of training and test samples are provided in the lower right corner.}
    \label{fig:deep5}
\end{figure}

The CNN-LSTM model is trained using weekly and bi-weekly flight images. Figure \ref{fig:deep5} demonstrates the performance of the deep learning model across six environments using 5 images on a weekly basis. According to the test performance, the $\text{r}^2$ between the ground truth and estimated relative maturity days are higher than 0.8 for all environments except environment 6. Environment 6 has achieved an $\text{r}^2$ of 0.5 on the test data. Having more outliers and a different UAV platform has resulted in a decrease in performance for this environment.

Environment 3 has the lowest mean absolute error (less than 1 day) amongst all environments. Environments 1, 4, and 5 have achieved a mean absolute error of fewer than 2 days, whereas environments 2 and 6 have a mean absolute error of almost 2 days.

\begin{figure}[!htb]
    \centering
    \includegraphics[width=12cm]{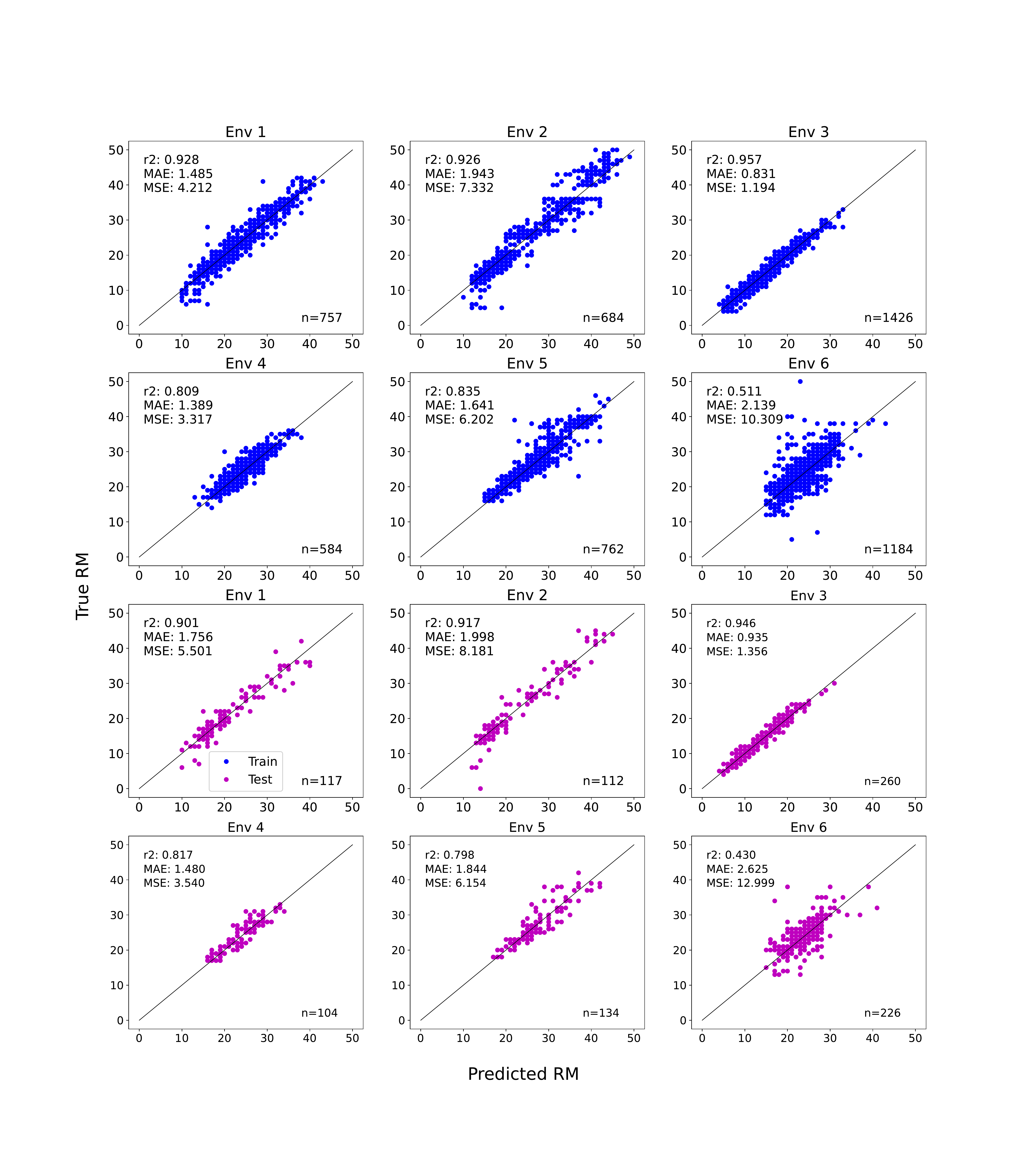}
    \caption{CNN-LSTM model's train (blue) and test (purple) performance across 6 environments for 3 images.}
    \label{fig:deep3}
\end{figure}

Figure \ref{fig:deep3} presents training and test results across six environments using 3 images. As expected, in most of the cases, $\text{r}^2$ has decreased with respect to the previous weekly model. Take, for example, environments 2, 3, 4, 5, and 6 have less $\text{r}^2$ compared to the previous model for both train and test sets. However, environment 1 has performed slightly better in terms of all three metrics, which can be explained by highly correlated images.

\subsubsection{Comparison}
As previously mentioned, the LOESS model depends upon setting a threshold value that is chosen through trial and error. Therefore, we estimate the LOESS model's performance using 9 different threshold values between 0.01 and 0.09. Detailed results are available in tables \ref{tab:loess5} and \ref{tab:loess3} (see appendix). The threshold value corresponding to the best performance differs across environments. Even in some cases, the optimal threshold differs with respect to the chosen metric (see Env2 and Env5 in table \ref{tab:loess3}). Using the LOESS model is difficult in practice due to the sensitivity of identifying the correct threshold. For soybean field trials placed around the entire Midwestern United States, identifying a different threshold value for each environment is not practical and makes scalable implementation difficult.

Table \ref{tab:compare} summarizes CNN-LSTM and LOESS performance for both weekly (using 5 images) and bi-weekly (using 3 images) analysis. It should be noted that the LOESS model is fitted to the time series of each plot separately, whereas the deep learning model fits a universal CNN-LSTM on training data and uses the test data for evaluation. 

In all cases, the CNN-LSTM model has achieved higher $\text{r}^2$ and lower MAE and MSE values, except for environment 1 where LOESS performs slightly better than the CNN-LSTM test performance. For environment 6, the LOESS model is not able to predict any variations on the response ($\text{r}^2<0$), whereas the CNN-LSTM model can predict almost 50\% of the variation. Furthermore, decreasing the number of input images from 5 to 3 has affected the performance of the LOESS model more than the CNN-LSTM model. 

\begin{table}[ht]
    \centering
    \begin{tabular}{|c|c|c|c|c||c|c|c|}
    \hline
        \multirow{3}{*}{Environment} & \multirow{3}{*}{Metric}& \multicolumn{3}{|c|}{Method (Weekly)} &  \multicolumn{3}{|c|}{Method (Bi-weekly)}\\
         \cline{3-8}
           &  & \multicolumn{2}{|c|}{CNN-LSTM} & LOESS  & \multicolumn{2}{|c|}{CNN-LSTM} & LOESS\\
           \cline{3-4}
           \cline{6-7}
            &  & Train & Test &  & Train & Test & \\
            \cline{1-8}
            \multirow{3}{*}{1} & $\text{r}^2$ & 0.925 & 0.889 & 0.906  & 0.928 & 0.901 & 0.851\\
                               & MAE & 1.486 & 1.796 & 1.760  & 1.485 & 1.756 & 2.152\\
                               & MSE & 4.394 & 6.201 & 5.562  & 4.212 & 5.501 & 8.637\\
            \hline
            \multirow{3}{*}{2} & $\text{r}^2$ & 0.939 & 0.925 & 0.788  & 0.926 & 0.917 & 0.746\\
                               & MAE & 1.761 & 2.012 & 3.068  & 1.943 & 1.998 & 3.520\\
                               & MSE & 6.079 & 7.439 & 21.078 & 7.332 & 8.181 & 25.317\\
            \hline
            \multirow{3}{*}{3} & $\text{r}^2$ & 0.963 & 0.949 & 0.940  & 0.957 & 0.946 & 0.908\\
                               & MAE & 0.779 & 0.901 & 0.921  & 0.831 & 0.935 & 1.217\\
                               & MSE & 1.032 & 1.283 & 1.644  & 1.194 & 1.356 & 2.523\\
            \hline
            \multirow{3}{*}{4} & $\text{r}^2$ & 0.839 & 0.819 & 0.719  & 0.809 & 0.817 & 0.313\\
                               & MAE & 1.259 & 1.494 & 1.653  & 1.389 & 1.480 & 2.852\\
                               & MSE & 2.795 & 3.509 & 4.955  & 3.317 & 3.540 & 812.125\\
            \hline
            \multirow{3}{*}{5} & $\text{r}^2$ & 0.870 & 0.878 & 0.759  & 0.835 & 0.798 & 0.616\\
                               & MAE & 1.493 & 1.505 & 2.155  & 1.641 & 1.844 & 3.086\\
                               & MSE & 4.905 & 3.722 & 8.843 & 6.202 & 6.154 & 14.071\\
            \hline
            \multirow{3}{*}{6} & $\text{r}^2$ & 0.537 & 0.506 & -0.001  & 0.511 & 0.430 & -0.160\\
                               & MAE & 2.095 & 2.503 & 3.377  & 2.139 & 2.625 & 3.913\\
                               & MSE & 9.760 & 11.270 & 21.362  & 10.309 & 12.999 & 24.770\\
            \hline
                           
    \end{tabular}
    \caption{Summary of weekly and bi-weekly analysis using CNN-LSTM and LOESS models. Performance of these two models with respect to three metrics, including $\text{r}^2$, MAE, and MSE, is demonstrated across six environments. For the CNN-LSTM model, both train and test performances are reported. For the LOESS model, a search grid is applied to find the optimal threshold, and then the best performance with respect to the corresponding metric is reported.}
    \label{tab:compare}
\end{table}

\section{Discussion and Implications}

For a commercial plant breeding organization, identifying the maturity date of soybeans is critical. When a soybean's relative maturity is not classified correctly, it affects harvest operations, can cause seed germination issues, affects advancement decisions, and contributes to a loss in genetic gain. With this approach, we have shown that our CNN-LSTM provides a robust approach to estimating a soybean's maturity date within $\pm$ 2 days of the actual maturity date without the ambiguity of the LOESS model. That is, we are robust to outliers and do not need to set arbitrary thresholds for prediction. Even using just three drone flights taken two weeks apart for predictions, we achieve high performance accuracy. 

The impact of this work demonstrates that drone flights do not need to be conducted every day and that if a drone flight can only be conducted in two week intervals, predictions will still be accurate. This will present itself as a cost saving measure, and no longer do operational specialists have to rush to the field to capture frequent drone flights or even to manually traverse the fields. 

Additionally, given the geographical considerations of environment locations, we also notice that some locations may need more drone flights than others, on average. For example, Environment 2 and Environment 6 will need at least 5 drone flights. Whereas Environments 1, 3, 4, and 5 having three drone flights is sufficient. This discrepancy can be attributed to differences in land topography, weather, or soil conditions. This inadvertently leads itself into other plant breeding questions that connect soybean genetics to the environment and farm management practices, all of which are important pieces to consider for soybean advancement.

Moreover, from the perspective of the plant breeder, knowing that an advancement decision can be made after taking three drone flights for a subset of locations is reassuring. Once a subset of locations has enough information for a confident decision, breeders can then focus their time gathering data about the rest of their field. Moreover, since an increase in drone flight only marginally improves accuracy for a subset of fields, a plant breeder can work with the operations team to create a schedule of drone flights that is flexible yet allows for enough data collection for the fields that are known to require more flights. Additionally, the need for fewer flights allows for more plots to be captured. As stated previously, it is common for a commercial organization to have 40,000 genetically different soybeans being grown at a single time, yet, our training data set only consists of 1,650 soybeans. Therefore with less demand on flight frequencies, organizations are now able to capture more data with fewer flights. That is, instead of simply having soybean relative maturity data for a small subset of varieties, organizations can now consider expanding drone programs to cover all their varieties. This leads to better decision making when determining the next soybean to advance to the next stage, and ultimately, be grown by farmers.

\section{Conclusions}

In this study, we demonstrated how an applied deep learning system could be used to help aid soybean breeding decisions. To achieve this, we developed an end-to-end framework to estimate the relative maturity of soybeans given the time series of UAV images.  A hybrid deep learning model is proposed to extract features from images and capture the temporal behavior of the time series data. Analyses are conducted for 6350 plots across six environments in Minnesota and Iowa in two growing seasons given two different flight frequencies (weekly and bi-weekly). The deep learning model can estimate the relative maturity days with less than 2 days of error for five environments among six total environments. The sixth environment has less than 3 days of error in prediction.

In predicting soybean relative maturity, color is the most important feature since soybean becomes mature when 95\% of the pods turn brown. Therefore, a successful model should be able to detect color properly. In a deep learning context, color is considered a simple feature that can be detected through earlier layers. This explains why we can achieve such a good performance using only four conventional layers in our proposed model.

To evaluate our estimations, we compared the deep learning results with a benchmark. The benchmark is a local regression model (LOESS), which models the greenness decay over time. We demonstrated the sensitivity of this method to its predefined threshold value. Although the local regression model is simple and fast, optimizing the threshold value can be analytically challenging and computationally expensive. This method maps each image to a single value (vegetation index) and does not keep all features of the images as what CNNs do. Furthermore, a separate model should be learned for the time series of each plot, which means the model cannot be generalized to new data. 

One of the advantages of the CNN-LSTM model is its robustness to data quality issues (e.g., dark and blur images). The other advantage of the deep learning model is its good performance in less frequent flights. Furthermore, our proposed method can be generalized to data from new environments.

With the accessibility of satellite imagery, future research in this area can expand on using satellite imagery to estimate soybean maturity. Moreover, given this framework, another research path would be to identify other soybean traits or traits from other crops that can be analyzed from drone imagery with the ultimate goal of improving decision making.

\section*{Conflicts of Interest}

The authors declare no conflicts of interest.

\section*{Acknowledgement}

This work was partially supported by Syngenta Company.

\newpage

\bibliographystyle{elsarticle-harv}
\bibliography{sample}

\clearpage
\section{Appendix}
\begin{table}[H]
    \centering
    \tabcolsep=0.11cm
    \begin{tabular}{|c|c|c|c|c|c|c|c|c|c|c|}
    \hline
        \multirow{2}{*}{Environment} & \multirow{2}{*}{Metric}& \multicolumn{9}{|c|}{Threshold} \\
         \cline{3-11}
           &  & 0.01 & 0.02 & 0.03 & 0.04 & 0.05 & 0.06 & 0.07 & 0.08 & 0.09\\
         \hline
         \multirow{3}{*}{1} & $\text{r}^2$ & 0.782 & 0.889 & \cellcolor[HTML]{FEF9E7}0.906 & 0.869 & 0.793 & 0.699 & 0.582  & 0.454 & 0.307\\
                            & MAE  & 2.835 & 1.926 & \cellcolor[HTML]{C0C0C0}1.760 & 2.145 & 2.810 & 3.526 & 4.276  & 0.454 & 5.738\\
                            & MSE   & 12.68 & 6.454 & \cellcolor[HTML]{EDBB99}5.562 & 7.628 & 12.037& 17.455& 24.273 & 31.717& 40.23\\
                                     \hline
         \multirow{3}{*}{2} & $\text{r}^2$ & 0.625  & \cellcolor[HTML]{FEF9E7}0.788 & 0.727  & 0.626 & 0.512 & 0.388 & 0.261 & 0.122 & -0.025\\
                            & MAE   & 4.710  & \cellcolor[HTML]{C0C0C0}3.068 & 3.868  & 4.863 & 5.770 & 6.647 & 7.457 & 8.255 & 9.020\\
                            & MSE   & 37.340 & \cellcolor[HTML]{EDBB99}21.078& 27.200 & 37.222& 48.534& 60.898& 73.598& 87.433& 102.028\\
                                     \hline
         \multirow{3}{*}{3} & $\text{r}^2$ & 0.130  & 0.557 & 0.763 & 0.874& 0.925 & \cellcolor[HTML]{FEF9E7}0.940 & 0.922& 0.883& 0.825\\
                            & MAE   & 4.558  & 3.149 & 2.164 & 1.439& 1.018 & \cellcolor[HTML]{C0C0C0}0.921 & 1.109& 1.463& 1.909\\
                            & MSE   & 23.964 & 12.194& 6.528 & 3.483& 2.060 & \cellcolor[HTML]{EDBB99}1.644 & 2.144& 3.219& 4.834\\
                                     \hline
         \multirow{3}{*}{4} & $\text{r}^2$ & -0.348 & 0.443 & 0.700 & \cellcolor[HTML]{FEF9E7}0.719 & 0.650 & 0.511 & 0.321 & 0.100 & -0.168\\
                            & MAE   & 3.828  & 2.353 & 1.711 & \cellcolor[HTML]{C0C0C0}1.653 & 1.840 & 2.266 & 2.798 & 3.391 & 4.00\\
                            & MSE   & 23.794 & 9.824 & 5.301 & \cellcolor[HTML]{EDBB99}4.955 & 6.183 & 8.632 & 11.978& 15.879& 20.605\\
                                     \hline
         \multirow{3}{*}{5} & $\text{r}^2$ & -0.079 & 0.563 & 0.739 & \cellcolor[HTML]{FEF9E7}0.759 & 0.688 & 0.561 & 0.395 & 0.212 & 0.002\\
                            & MAE   & 5.400  & 3.263 & 2.321 & \cellcolor[HTML]{C0C0C0}2.155 & 2.621 & 3.297 & 4.041 & 4.781 & 5.515\\
                            & MSE   & 39.520 & 16.018& 9.556 & \cellcolor[HTML]{EDBB99}8.843 & 11.444& 16.092& 22.144& 28.855& 36.539\\
                                     \hline
         \multirow{3}{*}{6} & $\text{r}^2$ & \cellcolor[HTML]{FEF9E7}-0.001 & -0.048 & -0.365 & -0.738 & -1.151 & -1.582 & -2.03 & -2.518 & -3.000\\
                            & MAE   & \cellcolor[HTML]{C0C0C0}3.377  & 3.650  & 4.387  & 5.184  & 5.933  & 6.647  & 7.333 & 8.011  & 8.635\\
                            & MSE   & \cellcolor[HTML]{EDBB99}21.362 & 22.367 & 29.144 & 37.096 & 45.931 & 55.128 & 64.688& 75.111 & 85.400\\
                            \hline
    \end{tabular}
    \caption{LOESS model performance across 6 environments (weekly analysis, using 5 images) with 9 different threshold values. For each metric, the best performance is highlighted.}
    \label{tab:loess5}
\end{table}

\begin{table}[H]
    \centering
    \tabcolsep=0.11cm
    \begin{tabular}{|c|c|c|c|c|c|c|c|c|c|c|}
    \hline
        \multirow{2}{*}{Environment} & \multirow{2}{*}{Metric}& \multicolumn{9}{|c|}{Threshold} \\
         \cline{3-11}
           &  & 0.01 & 0.02 & 0.03 & 0.04 & 0.05 & 0.06 & 0.07 & 0.08 & 0.09\\
         \hline
         \multirow{3}{*}{1} & $\text{r}^2$ &  0.277 & 0.616 & 0.777 & 0.845   & \cellcolor[HTML]{FEF9E7}0.851 &  0.806 & 0.722  & 0.605 & 0.45\\
                            & MAE  & 5.403 & 3.727 & 2.675   & 2.196   & \cellcolor[HTML]{C0C0C0}2.152 & 2.545 & 3.197  & 4.003 & 4.888\\
                            & MSE   & 41.943 & 22.299 & 12.966 & 9.026 & \cellcolor[HTML]{EDBB99}8.637& 11.279& 16.11 & 22.949& 31.938\\
                                     \hline
         \multirow{3}{*}{2} & $\text{r}^2$ & 0.317  & \cellcolor[HTML]{FEF9E7}0.746 & 0.745  & 0.667 & 0.549 & 0.413 & 0.267 & 0.102 & -0.082\\
                            & MAE   & 6.815  & 3.618 & \cellcolor[HTML]{C0C0C0}3.520  & 4.180 & 5.131 & 6.141 & 7.078 & 8.044 & 8.989\\
                            & MSE   & 68.011 & \cellcolor[HTML]{EDBB99}25.317& 25.359 & 33.139& 44.874& 58.417& 72.960& 89.358& 107.68\\
                                     \hline
         \multirow{3}{*}{3} & $\text{r}^2$ & -0.910  & -0.185 & 0.247 & 0.539& 0.728 & 0.839& 0.897& \cellcolor[HTML]{FEF9E7}0.908& 0.887\\
                            & MAE   & 6.830  & 5.265 & 4.073 & 3.076  & 2.258 & 1.680 & 1.316& \cellcolor[HTML]{C0C0C0}1.217& 1.353\\
                            & MSE   & 52.625 & 32.647& 20.751 & 12.693& 7.500 & 4.437 & 2.85& \cellcolor[HTML]{EDBB99}2.523& 3.113\\
                                     \hline
         \multirow{3}{*}{4} & $\text{r}^2$ & -2.605 & -1.242 & -0.419 & 0.025 & 0.261 & \cellcolor[HTML]{FEF9E7}0.313   & 0.217 & -0.021 &  -0.380\\
                            & MAE   & 7.096  & 5.430 & 4.180 & 3.458   & 2.988 & \cellcolor[HTML]{C0C0C0}2.852   & 2.964 & 3.352 & 3.958\\
                            & MSE   & 63.61 & 39.564 & 25.041 & 17.202 & 13.038 & \cellcolor[HTML]{EDBB99}12.125 & 13.821& 18.02 & 24.344\\
                                     \hline
         \multirow{3}{*}{5} & $\text{r}^2$ & -0.594 & -0.005 & 0.347 & 0.542   & \cellcolor[HTML]{FEF9E7}0.616 & 0.591 & 0.489 & 0.331 & 0.099\\
                            & MAE   & 6.783  & 5.390  & 4.331 & 3.557   & 3.158 & \cellcolor[HTML]{C0C0C0}3.086 & 3.355 & 3.940 & 4.821\\
                            & MSE   & 58.377 & 36.796 & 23.903 & 16.767 & \cellcolor[HTML]{EDBB99}14.071& 14.985& 18.723& 24.511& 33.000\\
                                     \hline
         \multirow{3}{*}{6} & $\text{r}^2$ & -0.609 & \cellcolor[HTML]{FEF9E7}-0.160 & -0.242 & -0.537 & -0.954 & -1.457 & -2.029 & -2.673 & -3.378\\
                            & MAE   & 4.595  & \cellcolor[HTML]{C0C0C0}3.913  & 4.086  & 4.649  & 5.440  & 6.315  & 7.227 & 8.157  & 9.055\\
                            & MSE   & 34.360 & \cellcolor[HTML]{EDBB99}24.770 & 26.527 & 32.812 & 41.718 & 52.450 & 64.668& 78.418 & 93.460\\
                            \hline
    \end{tabular}
    \caption{LOESS model performance across 6 environments (bi-weekly analysis, using 3 images) with 9 different threshold values. For each metric, the best performance is highlighted.}
    \label{tab:loess3}
\end{table}

\end{document}